\documentclass{article}
\usepackage{spconf,amsmath,graphicx}
\usepackage{listings}
\usepackage{protobuf/lang}  % include language definition for protobuf
\usepackage{protobuf/style} % include custom style for proto declarations.
\usepackage[T1]{fontenc}

\title{Transformer-based Language Modeling and decoding
for Conversational Speech Recognition}
\name{Kareem Nassar}
\address{Columbia University in the City of New York\\
\texttt{kan2140@columbia.edu}}
\begin{document}
%\ninept
%
\maketitle
\begin{abstract}
We propose a way to use a transformer-based language model in conversational speech recognition. Specifically, we focus on decoding efficiently in a weighted finite-state transducer framework. We showcase an approach to lattice re-scoring that allows for longer range history captured by a transfomer-based language model and takes advantage of a transformer's ability to avoid computing sequentially.
\end{abstract}
\section{Introduction}
\label{sec:intro}
\begin{figure}[ht]
    \centering
    \includegraphics[width=8cm]{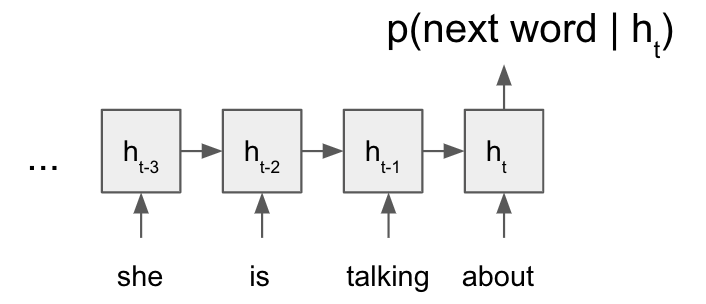}
    \caption{A recurrent neural network predicting the probability of the next word using hidden states}
    \label{fig:rnn}
\end{figure}

\begin{figure}[ht]
    \centering
    \includegraphics[width=8cm]{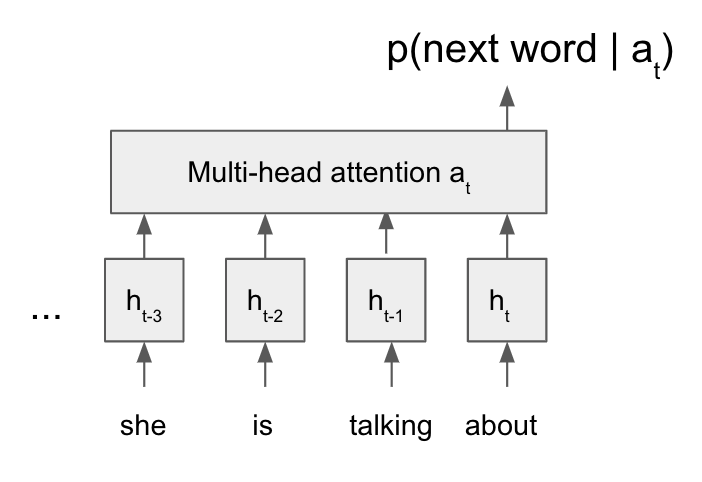}
    \caption{A simplified view of the transformer architecture. No recurrence, just attention.}
    \label{fig:transformer}
\end{figure}

In conversational speech, individual utterances may reference context from previous utterances. If certain topics or words have been mentioned in the past, related words are likely to be used. In automatic speech recognition, language models are responsible for capturing the probabilities of words likely to be uttered given past words. Traditionally, these probabilities are captured in an n-gram language model. For example, in a trigram language model, we would store a mapping of all 3-word combinations found in our training corpus, along with the probabilities of the third word following the previous two words. The past "history" that a trigram language model would capture is limited to two words. It is also limited in its ability to capture semantics. With the advent of more computational power, neural-based methods for language modeling, like the Recurrent Neural Network (RNN) architecture have become possible. In neural architectures, words are typically represented as word embeddings, n-dimensional vectors that attempt to capture the semantics in the latent space. Due to its recurrent set-up, the RNN-based language model is able to encapsulate previous word embeddings in its hidden state (see Figure \ref{fig:rnn}).
The hidden state at each step represents a single value computed from a series of matrix operations on current word embedding and the hidden state from the previous step. However, with a long sequence, RNNs on their own suffer from vanishing gradients, Long Short-Term Memory (LSTM) units remedy this by introducing additional operations for computing each hidden state \cite{GRAD}. Despite their name, they fall short of capturing word dependencies past 200 words and focus on past 50 words more heavily \cite{LSTM_issues}. The transformer architecture, however, does not suffer from this problem. It does away with recurrence in favor of an attention mechanism \cite{ATTENTION}. The attention mechanism allows the network to learn a weighted average to determine which words are important at at each position. The network can capture much longer dependencies as all previous words are encoded and passed into the multi-head attention layers in parallel (see Figure \ref{fig:transformer}). The multi-head attention layer learns to reference words from the beginning of a very long context. This is a desirable attribute that we'd like to apply to conversational speech recognition, however, there is one problem: many transformer-based architectures take in fixed size input \cite{BERT}. Any change to the input, whether it be adding another word, or changing the last word in the input would require recomputing all values of the network. This can be prohibitive in the case of speech recognition as it would require the re-computation of all word inputs from the beginning of our speech context for every new utterance that we attempt to re-score. The transformer-XL architecture solves this by introducing a segment-level recurrence mechanism \cite{Transformer_xl}. In this work we use the Kaldi framework \cite{kaldi} and a transfomer-XL architecture to do efficient lattice re-scoring. For each new utterance, we cache the segment-level embeddings for use in future utterance lattice re-scoring.

\section{Past Work}
\label{sec:past}
Typically, due to Kaldi’s use of weighted finite state transducers, the fast decoding techniques require that a language model is expanded as a finite state transducer \cite{wfst}. This is difficult to do with an RNNLM as it requires approximating the RNN's probability distribution and converting into the a deterministic FST form. There have been attempts to sample an RNNLM to produce an equivalent FST, however the accuracy performance was equivalent to that of a bigram language model \cite{rnn_to_fst}.
Because of this, first-pass decoding is done using an n-gram language model. After first-pass decoding with an n-gram language model, a second-pass re-scoring is done using an RNNLM. However, this can be slow, so n-gram approximation is used, but this limits the history that an RNN has available to it. \cite{pruned_rnn} proposes a pruned approach which allows increasing the n-gram approximation to more n-grams but without degrading performance.
In other works to handle long range dependencies, \cite{rnn_adaptation} use an RNNLM with a "conversation cache" and DNN-based adaptation technique. The "conversations cache" is a count of seen unigrams. The cache is used to modify unigram priors before RNNLM re-scoring.
% Your proposed method for solving the problem
% Proposed data that will be used for your project and where you are getting the data
% Expected results
 
\section{Approach}
\label{sec:approach}
We use XLNet \cite{xlnet} to attempt to capture long range language dependencies. At the time of this writing, XLNet provides the best accuracy for many downstream tasks that require language modeling pre-training, including question-answering,  text classification, and other natural language understanding tasks. We also attempt to take advantage of a transformer's parallel properties to make some performance optimizations when re-scoring our lattices.
\subsection{Why XLNet over BERT?}
\label{sec:xlnet}
\begin{figure}[ht]
    \centering
    \includegraphics[width=6cm]{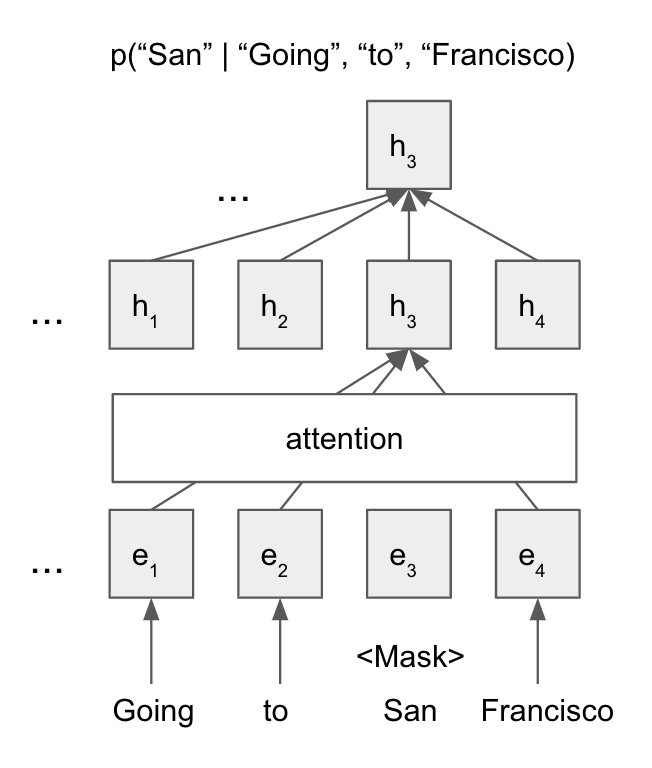}
    \caption{BERT attempts to predict the masked word use both left and right contexts. During training, a certain percentage of words are masked for use in prediction. If both "San" and "Francisco" were masked, BERT would not be able to use information when decoding one of the words to help in decoding the other.}
    \label{fig:bert_masking}
\end{figure}

\begin{figure}[ht]
    \centering
    \includegraphics[width=6cm]{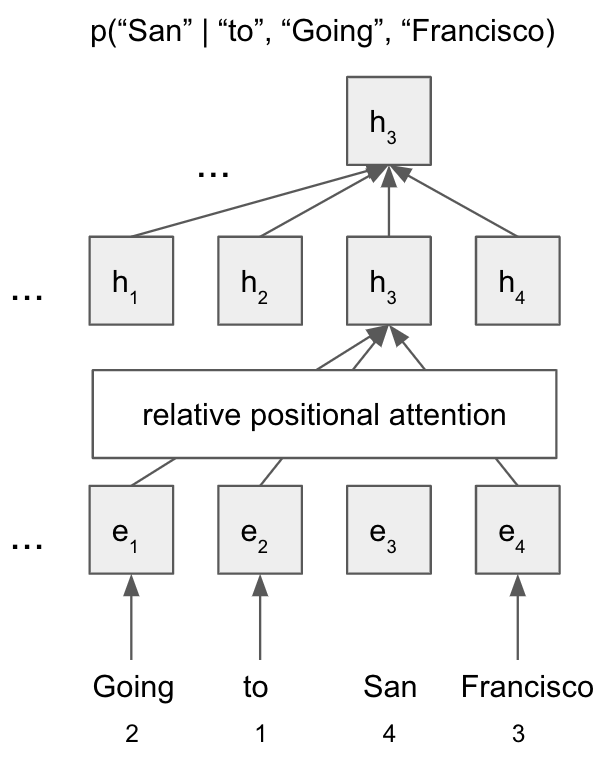}
    \caption{In XLNet training, every sample can have a different permutation ordering. Instead of passing an explicit masking token like BERT, the permutation ordering dictates what is visible to each token's hidden states across the network. Relative positional encoding preserve information about the original sequence ordering. }
    \label{fig:xlnet_permutation}
\end{figure}

XLNet is a generalized auto-regressive model that can be used for language modeling based on the transformer-XL architecture \cite{xlnet}. This means that the outputs of XLNet depend strictly on the previous outputs. This is different from other state-of-the-art language models like BERT (Bidirectional Encoder Representations from Transformers) which rely on conditioning the probabilities given surrounding words. In BERT, the model tries to predict a masked word by looking at all surrounding unmasked words (figure \ref{fig:bert_masking}).
The concept of masking the input introduces a few disadvantages mentioned in \cite{xlnet}. Firstly, a masked token is rarely seen for most subsequent language modeling tasks, so there tends to be a discrepancy between the "pre-taining" step and the "fine-tuning" step. Typically, in the "fine-tuning" step, BERT is adapted to attempt tasks like question-answering. Secondly, BERT does not use information from one decoded masked token to help in decoding another masked token. In other words, all masked tokens are assumed to be independent. This is necessary in BERT, because there is a strict separation between unmasked and masked tokens, as the masked tokens will be predicted. In XLNet, however, the separation is a directional one: anything to the left of the word that is attempting to be predicted is fair game (during training, words are reordered to get the benefit of surrounding context, but conceptually, orders are seen from left to right for the factorization order). This lends itself well to decoding in speech recognition as we typically re-score a lattice from left to right (assuming you are visualizing a lattice in english), while we prune low scoring results. This does not mean, however, that the encodings for the words do not capture context from surrounding words. With permutation language modeling, during training, the ordering of previous tokens can be modified (figure \ref{fig:xlnet_permutation}).
With permutation language modeling, the network is trained on a regiment of random order word sequences. For example, the phrase: "I am going to San Francisco to watch the Warriors play basketball" could be used to train XLNet by selecting a random factorization order, which is the order in which we will decode the network "see" the tokens. Since this is accomplished by passing an attention mask, which controls which positions are allowed to attend to which other positions, we don't give up any parallelism. Then we select a pivot index, say 6 corresponding to the word "to." All words preceding the pivot would be used to predict words succeeding the pivot. So we may try to predict the p("to"| "I", "am", "going", "to", "San", "Francisco") and p("to" | "San", "am", "I", "to", "going", "Francisco"), among all other permutations of the preceding words (see figure \ref{fig:xlnet_permutation}). Although this random ordering seems jarring, the original word sequence orderings are still preserved through "relative positional embeddings." XLNet maintains as an input, the relative word position distance from the word that is being predicted from its inputs. This means that a random permutation of previous words can help the network learn an embedding from the surrounding words while it preserves information related to the ordering of those words.
\begin{figure}[ht]
    \centering
    \includegraphics[width=6cm]{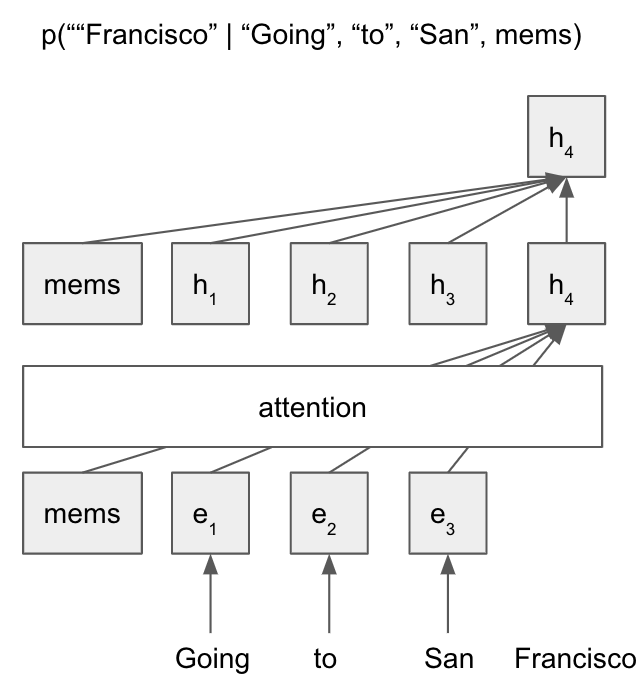}
    \caption{The transformer-XL, the architecture that XLNet is based on, allows for caching of hidden states from previous outputs. These states can be used as inputs into the attention layers.}
    \label{fig:xlnet_mems}
\end{figure}
Another feature of XLNet, is that it allows the exporting of its internal hidden states, which can be passed in on subsequent calls. This provides a recurrence mechanism that can help capture long range dependencies (see figure \ref{fig:xlnet_mems}). The segment-level memory units (mems) allow us to capture longer range dependencies in language without recomputing all previous hidden states. There are large performance improvements for longer attention lengths as well \cite{Transformer_xl}.
\subsection{Decoding and Re-scoring in Kaldi}
\label{sec:decoding}
\begin{figure}[ht]
    \centering
    \includegraphics[width=6cm]{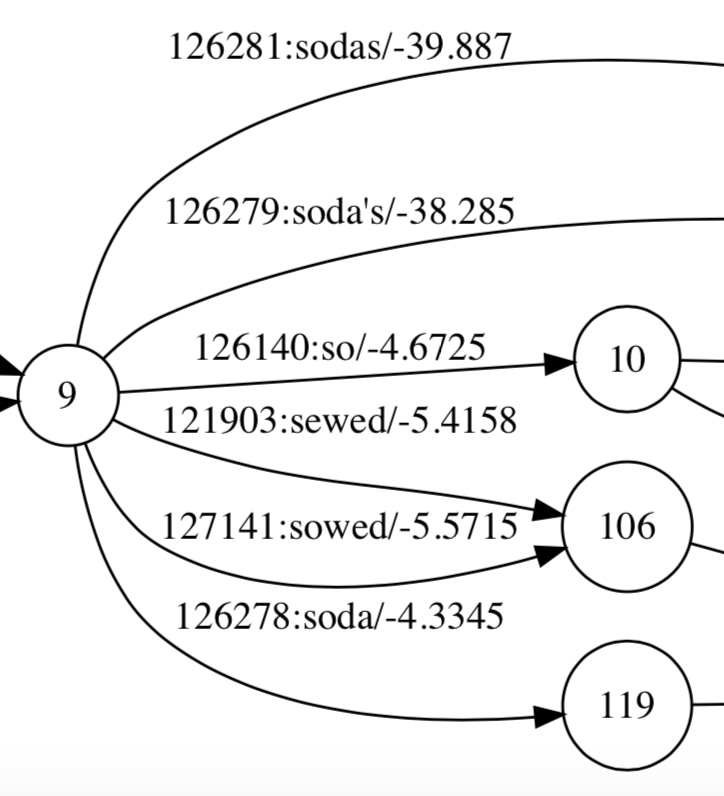}
    \caption{This is an example of a lattice. "So does" or "sodas"? Each arc can represent a word and a log probability that combines the acoustic and language modeling scores. The best path for the lattice represents the best scoring sentence decoded for this utterance.}
    \label{fig:lattice_fst}
\end{figure}

Weighted finite state transducers (WFSTs or FSTs) provide a structured way to perform decoding. They represent a directed graph where each arc can represent probabilities of various state transitions. FSTs can represent hidden Markov models (HMMs), context dependency models, lexicons (pronunciation dictionaries), and grammars (n-gram language models). Kaldi uses FSTs heavily, and combines all of the previously mentioned components into a single composed FST (HCLG.fst) \cite{wfst}. During decoding, an acoustic model is used to produce log probabilities of various phonemes, and the HCLG.fst will be used to map the outputs of the acoustic model to HMM state transitions to context-dependent phones (H.fst) to context-independent phones (C.fst) to pronunciations of a single word (L.fst) to subsquent words (G.fst). The resulting output is a lattice which can also be represented as an FST (see figure \ref{fig:lattice_fst}). The lattice contains the combined acoustic model and language model probabilities for each arc (see figure \ref{fig:lattice_fst}).
The language modeling probabilities represent the log probability of the next state given the previous n-grams from the path leading up to that state.
Typically, for larger language models, a separate re-scoring step occurs where the arcs in a lattice are re-scored by subtracting the initial language modeling score that was given to a particular arc and inserting the new score from the larger language model. In a pruned approach, only a subset of the arcs are re-scored. Using a set of heuristics \cite{pruned_rnn}, arcs that require re-scoring are added to a priority queue and only the most promising paths are explored. In deterministic approaches, all possible paths are explored and re-scored. 
The re-scoring lattice operation takes the form of an FST composition. In FST composition \cite{openfst} two FSTs are combined to form an FST with the combined weights for the arcs that "exist" in both.

\section{Experiments and Results}
\label{sec:exp}
\begin{figure}[ht]
    \centering
    \includegraphics[width=8cm]{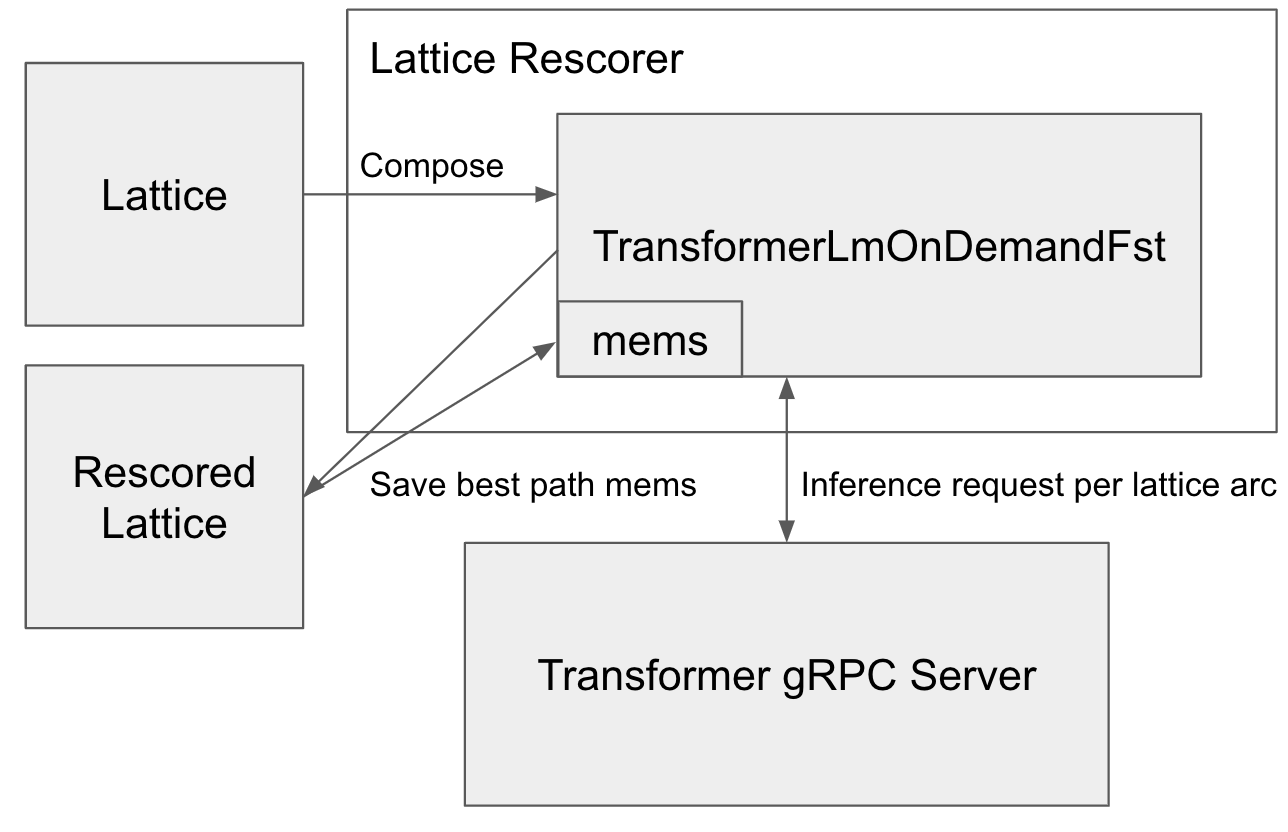}
    \caption{Our architecture for lattice re-scoring: TransformerLmOnDemandFst is composed with the lattice to re-score each lattice arc. Each inference request is sent to our gRPC server. After re-scoring the lattice, the best path is passed to the transformer to get its memory states, those are cached in the TransformerLmOnDemandFst for rescoring future sequential utterances.}
    \label{fig:rescoring_flow}
\end{figure}

All experiments were conducted using the TED-LIUM3 dataset \cite{ted_lium3}. TED-LIUM is an English speech recognition training corpus from TED talks. This data-set was chosen due to its topical nature, usually in the form of 15 minutes or more worth of speech where the speaker is discussing a particular topic. Our TED-LIUM dataset contains a training set of 248 hours of speech with aligned transcription, with approximately 2 hours of development and 3 hours of test. An acoustic model and n-gram language model are trained to provide a baseline word-error rate.
We use a library that contains a pre-trained version of XLNet, an implementation of the transformer-XL architecture \cite{hugging}. The model is fairly large with 110M parameters. It was previously trained on BooksCorpus \cite{book_corpus} and English Wikipedia which have 13GB of plain text combined \cite{xlnet}. We run a transfer learning step using PyTorch on the TED-LIUM dataset. We implement a gRPC server that can run inference on our model over the local network. In Kaldi, we implement a DeterministicOnDemandFst that calls into our exported model. Our DeterministicOnDemandFst maps kaldi word symbol identifiers to tokens that our model understands and vice-versa. When re-scoring a lattice, we remove the first-pass FST values and compose our DeterministicOnDemandFst (see figure \ref{fig:rescoring_flow}). The segment embedding for the best path after lattice re-scoring is cached and passed as inputs for re-scoring future lattices from the same speech context. We compare this technique to first-pass decoding lattices (no-rescoring) and to re-scoring with an RNNLM trained directly on the TED-LIUM data-set.
\subsection{XLNet gRPC Inference Server}
For ease of integration, we implement a gRPC server for access over the local network. gRPC provides a high-performance, cross-language way to send data across processes and over the network. Our implementation of a gRPC server takes in a sequence of words and returns the log probability for the next word.

\begin{lstlisting}[language=protobuf3,style=protobuf]
syntax = "proto3";

package edu.columbia.reco;

service Transformer {
  rpc GetLogProb(GetLogProbRequest) returns (GetLogProbResponse) {}
  rpc GetLogProbBatch(GetLogProbBatchRequest) returns (GetLogProbBatchResponse) {}
}

message GetLogProbRequest {
  repeated string previous_ngrams = 1;
  int64 mems_id = 2;
  string word = 3;
}

message GetLogProbBatchRequest {
  repeated GetLogProbRequest requests = 1;
  int64 common_mems_id = 2;
}

message GetLogProbResponse {
  float log_prob = 1;
  int64 mems_id = 2;
}

message GetLogProbBatchResponse {
  repeated GetLogProbResponse responses = 1;
}
\end{lstlisting}
There are two methods that can be invoked on our gRPC server: GetLogProb and GetLogProbBatch.  GetLogProb can take a single sequence of words and returns the log probability of the next word. It will also return a "mems\_id" which can be used on subsequent requests to reference the cached hidden memory states on the gRPC server. GetLogProbBatch can take a batch of requests and execute them in a single batch on the model. Words are tokenized to map to the word IDs that the XLNet model understand. Also, all sequence lengths in the batch are expected to be the same, so a padding token is added to pad all sequences to the maximum length sequence of the batch. A "common\_mems\_id" can be be passed with a batch request to load the cached hidden memory states as part of the batch inference call into the model. The memory states are repeated to match the dimensions of the batch. After each sequential lattice is re-scored, its best path is passed as sequence of words to GetLogProb and predicting the probability of an end of sentence token. The returned "mems\_id" is saved.

\subsection{Fine-tuning Language Model}
We fine-tune the XLNet base-cased model for 20 epochs on the TED-LIUM training set. First we concatenate all of the TED-LIUM training transcripts, preserving the order of the utterances. This step is important because we want to be able to condition our language models on very long histories of words (more than 100) so we need to be sure that contiguous training text belongs to the same TED talk. The training text is then tokenized using XLNet's word dictionary. Our training examples consist of blocks of 512 XLNet word IDs. We use an Adam optimizer with a learning rate of 5e-6. We did not attempt to different factorization orders (figure \ref{fig:xlnet_permutation}), only the natural factorization order was used. The cross entropy losses against all of the word positions were back-propagated for experiments where we had no target mapping. Other training runs for 10 and 4 word blocks were also ran.
We changed the random sampling to sequential sampling during training to allow for memory blocks to be cached from previous examples. We also experimented with a single permutation mask for the next token in order to mimic inference time where we are attempted to predict a last word that all other word positions and hidden states cannot attend to.
There are a few discrepancies between the pre-trained XLNet and our use-case. All of our words are lower cased and our corpus contains no punctuation. Since the pre-trained XLNet differentiates between upper and lower cased words and has punctuation tokens, it is likely that it relies on them for deriving semantics.  XLNet's pre-trained vocabulary is also much larger than TED-LIUM's. 
\subsection{Performance Optimizations}
\label{sec:perf}
\begin{figure}[ht]
    \centering
    \includegraphics[width=8cm]{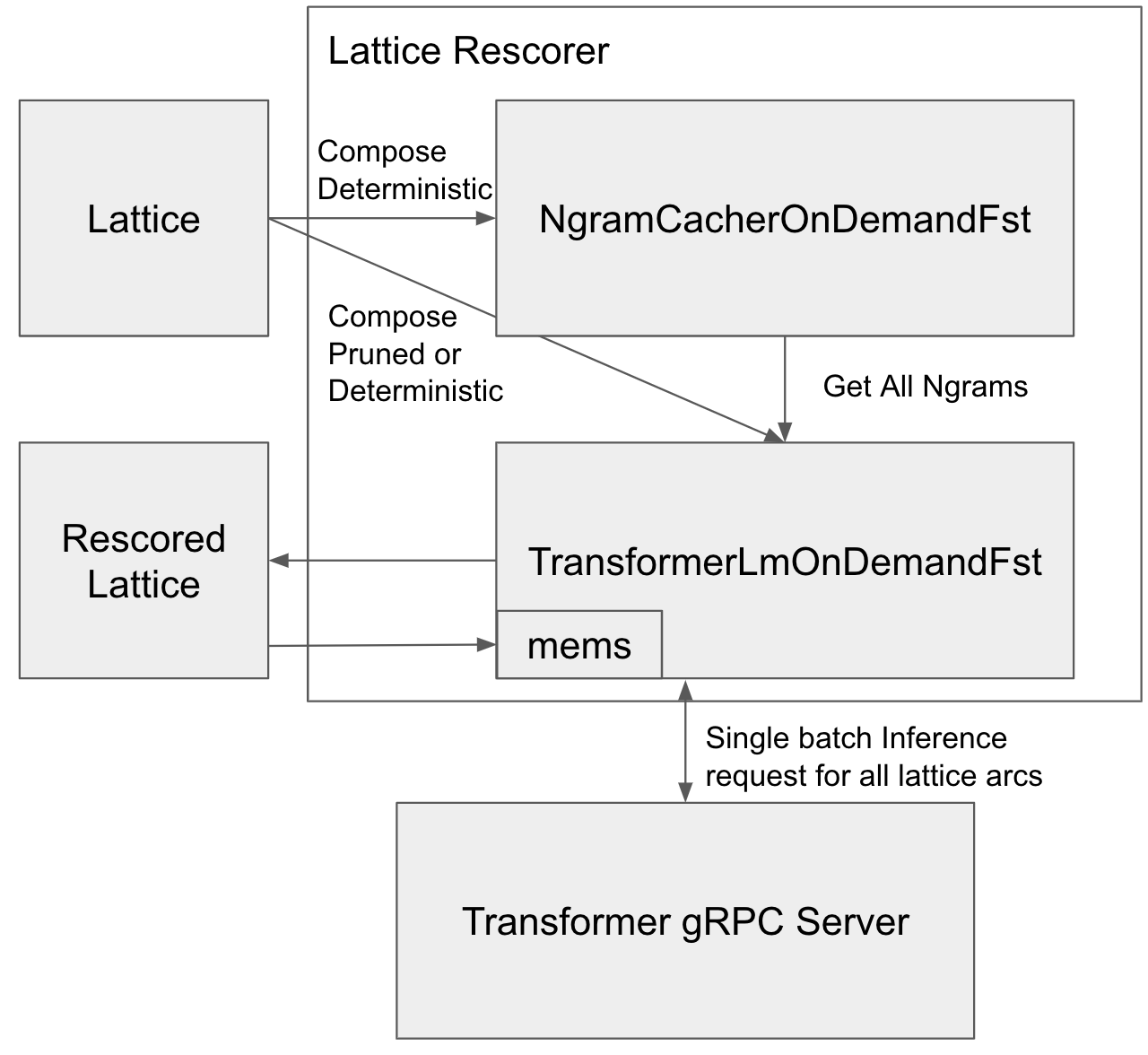}
    \caption{An improvement to our lattice rescoring flow involves the NgramCacherOnDemandFst to retrieves all n-grams in the lattice. All n-grams are passed to the TransformerLmOnDemandFst for a single batch request to the transformer server.}
    \label{fig:rescoring_flow2}
\end{figure}
Typically with an RNNLM approach, we must have cached the hidden states of previous time steps before running inference for the next FST state. This is improved upon with our transformer LM approach, as we can compute the log probability of the next unigram with a single inference pass. There is no need to compute the hidden state for each additional word in sequence. The problem, however, is that the transformer based language model (110M parameters for the one used on our experiments) is much larger than the RNNLM (15M parameters) that is referenced in the TED-LIUM recipe in Kaldi. This makes running the model a much more expensive prospect. For that reason, we must take a different approach to running our model. Because we are not constrained by running our models in sequence, we can batch all operations for execution in a single-pass, if the GPU memory permits it (see figure \ref{fig:rescoring_flow2}). 
\begin{figure}[ht]
    \centering
    \includegraphics[width=8cm]{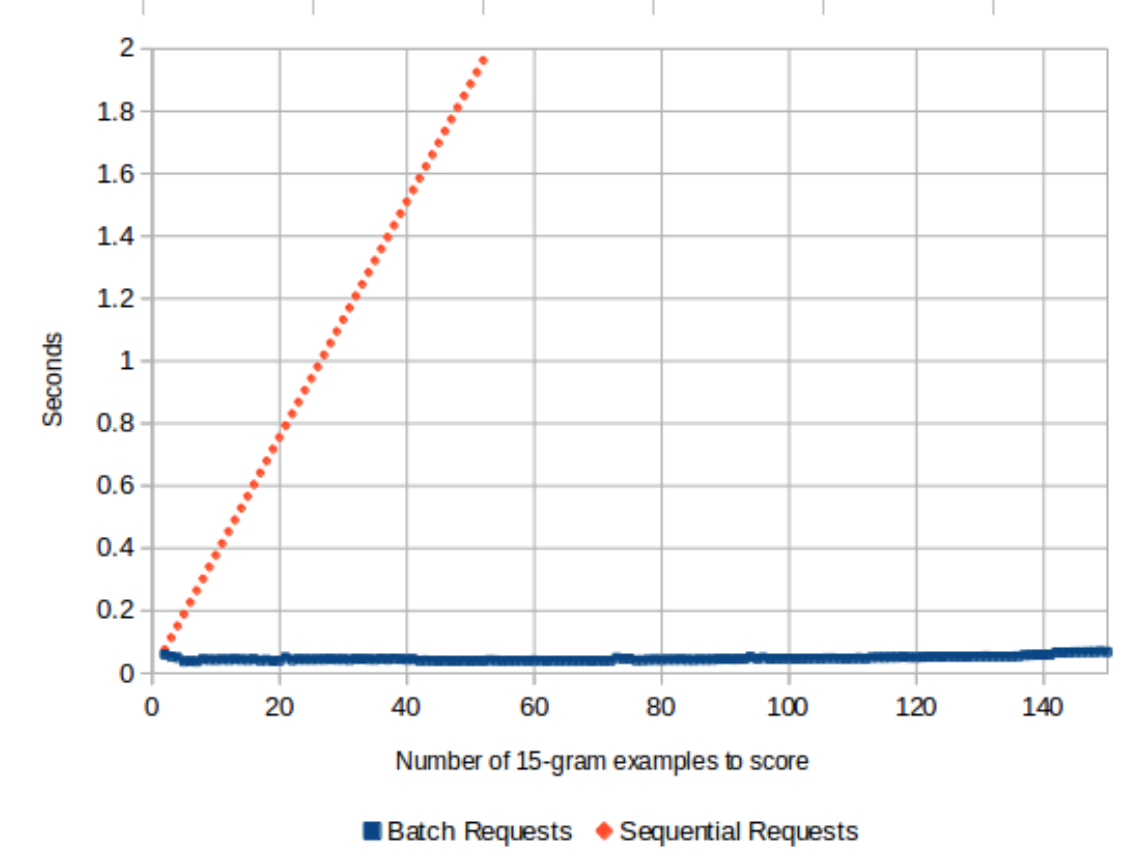}
    \caption{This is a graph showing the incremental cost of adding a request to the batch.}
    \label{fig:perf_res}
\end{figure}
See the graph in figure \ref{fig:perf_res} which shows that the incremental cost of adding an extra call to the batch operation is negligible (see figure \ref{fig:perf_res}).
This allows us to experiment with different techniques, for one, we can relax our pruning parameters as there is very little additional cost to run a deterministic approach where we score every arc in the lattice.

\subsection{Results}
\label{sec:res}
%This should include details of what data is being used and where the data was obtained.  Details about the nature of the data should be given.  The whole experimental set up, including training set, dev set, test set, sizes, quality, and many other details of the manipulation of the data, features, backend, recognition, etc.  This should be followed by quantitative results in the form of a table, graph, or both, with comparison with the state of the art, preferably on the same dataset.
\begin{table}[]
    \centering
 \begin{tabular}{c c c} 
 \hline
 Model & Dev WER & Test WER\\ [0.5ex] 
 \hline
 No-rescoring & 8.75 & 8.77 \\ 
 \hline
 Base XLNet 4-gram, no mems & 8.54 & 8.73 \\
 \hline
 Base XLNet 4-gram, with mems & 8.46 & 8.82 \\
 \hline
 Fine-tuned XLNet 4-gram & 8.32 & 8.47 \\
 \hline
 Fine-tuned XLNet 10-gram & 8.29 & 8.44 \\
 \hline
 RNNLM 4-gram & 6.77 & 7.25  \\
 \hline
 Lattice Oracle Best Path & 1.81 & 1.69  \\
 \hline
\end{tabular}
    \caption{The pre-trained and fine-tuned transformer models make improvements on the lattice, but they're still not as good as the pre-trained RNNLM for TED-LIUM. This is most likely due to using a very large XLNet, limited GPU resources, and only 25MB worth of TED-LIUM text.}
    \label{tab:res}
\end{table}
In order to motivate the problem, we measure the oracle word-error rate which gives us the path with the minimum word error rate found within each lattice. The oracle word error rate for the test set was found to be 1.70\%. If we were to flawlessly re-score a lattice, we could, in theory, achieve this word error rate. In the lattice some very good answers exist. However, our results in table \ref{tab:res} show how difficult it is to make a dent in the WER with such a large XLNet model. The RNNLM still gives a much better score. We suspect that this is due to a few things: firstly, the XLNet is 110M parameters and was trained on approximately 13GB of text compared to 25MB worth of text for TED-LIUM. Given the size of the model, and the fact that it was pre-trained on 512 TPUs \cite{xlnet}, we expect that training for 20 epochs on TED-LIUM's text is not enough to overcome the differences between written text and conversational speech. Without fine-tuning, adding memory seems to have an adverse effect on the test set.

\section{Conclusion}
\label{sec:expected}
We proposed a way to use a transformer-based language model in conversational speech recognition. We showed that XLNet-based language model with 110M parameters is slower in lattice re-scoring than the RNNLM, but there is promise when it comes to the ability of the transformer architecture to run inference in parallel. For a fairer comparison, a much smaller derivative of XLNet should be used. The transformer also showed a 4\% relative improvement on the TED-LIUM when transfer learning and running a 10-gram approximation on the lattice. Given a smaller model and more training time, we expect to see small XLNets to be used for lattice re-scoring soon.

\section{Future Improvements}
\label{sec:future}
Future improvements include reducing the batch sizes that are sent to our inference server by caching common n-gram prefixes. We also experimented fine-tuning the language models with memory hidden states but did not have time to compile the results. A much smaller network based on the XLNet archicture would allow for easier fine-tuning and inference. It also may be easier to test using the pre-computed word embedding matrix from TED-LIUM's RNNLM. That would help with reducing the size of the model overall and training time since we could avoid using the model's internal embedding lookup matrix. We would also like to explore scoring lattices in both directions and re-scoring the n-best outputs directly. We also wish to experiment with expanding the lattice with unique n-gram paths first before re-scoring to reduce the effect of multiple states affecting the path.
\bibliographystyle{IEEEbib}
\bibliography{strings,refs}

\end{document}